\documentclass[copyright,creativecommons]{eptcs}
\usepackage{comment}
\usepackage{mathptmx}
\usepackage{amsmath}
\usepackage{hyperref}
\usepackage{xspace}
\usepackage{todonotes}
\usepackage{listings}
\usepackage{enumitem}
\usepackage{titlesec}
\usepackage{etoolbox}

\makeatletter
\patchcmd{\ttlh@hang}{\parindent\z@}{\parindent\z@\leavevmode}{}{}
\patchcmd{\ttlh@hang}{\noindent}{}{}{}
\makeatother
\usepackage{caption}
\setlist{nolistsep,itemsep=0pt}

\def\myparagraph#1{
	\smallskip
	\noindent{\em #1}:~}

\titlespacing*{\section}{0pt}{0.5\baselineskip}{0.2\baselineskip}
\titlespacing*{\subsection}{0pt}{0.5\baselineskip}{0.2\baselineskip}

\def\class#1{{\hbox{{\sc #1\/}}}}
\def\role#1{{\hbox{{\it #1\/}}}}
\def\action#1{{\hbox{{\it #1\/}}}}
\def\verb#1{\hbox{\it #1}\xspace}
\def\sort#1{{\hbox{{\it #1\/}}}}

\def\attribute#1{{\hbox{{\it #1\/}}}}
\def\fluent#1{{\hbox{{\tt #1\/}}}}

\def\ajin{\mathit{ajin}}
\def\ajout{\mathit{ajout}}
\def\asin{\mathit{asin}}
\def\asout{\mathit{asout}}

\def\js{{\em JS}\xspace}

\def\lth{{\sc lth}\xspace}
\def\propbank{{\sc PropBank}\xspace}
\def\verbnet{{\sc VerbNet}\xspace}
\def\semlink{{\sc SemLink}\xspace}
\def\wordnet{{\sc WordNet}\xspace}
\def\framenet{{\sc FrameNet}\xspace}
\def\corealmlib{{\sc CoreALMLib}\xspace}
\def\corecalmlib{{\sc CoreCALMLib}\xspace}
\def\vnclasslib{{\sc vn\_class\_library}\xspace}
\def\strips{{\sc strips}\xspace}

\def\core{{\sc coreNLP}\xspace}
\def\babi{{ bAbI}\xspace}
\def\babiplus{{ bAbI+}\xspace}
\def\propara{{\sc ProPara}\xspace}
\def\calm{{\sc calm}\xspace}
\def\text2alm{{\sc text2alm}\xspace}
\def\textdrs{{\sc text2drs}\xspace}
\def\drsalm{{\sc drs2alm}\xspace}
\def\sphinx{{\sc sphinx}\xspace}
\usepackage[mathscr]{euscript}
\def\alm{\hbox{${\mathscr{ALM}}$}\xspace}

\def\beq{\begin{equation}}
\def\eeq#1{\label{#1}\end{equation}}
\def\ba{\begin{array}}
	\def\ea{\end{array}}

\newcommand\bcmdtab{\noindent\bgroup\tabcolsep=0pt%
  \begin{tabular}{@{}p{10pc}@{}p{20pc}@{}}}
\newcommand\ecmdtab{\end{tabular}\egroup}

\title{Information Extraction Tool \text2alm: 
        	From Narratives to Action Language System Descriptions
	\thanks{We would  like to thank Parvathi Chundi, Nicholas Hippen, Brian Hodges, Joseph Meyer, Gang Ling, and Ryan Schuetzler for their valuable feedback. We appreciate the insights from Michael Gelfond, Daniela Inclezan,  Edward Wertz, and Yuanlin Zhang on their work on language \alm, the \corealmlib library, and system \calm.}}

\author{%
  Craig Olson ~~~~ Yuliya Lierler
  \institute{University of Nebraska Omaha\\
    6001 Dodge St, Omaha, NE 68182, USA}
          \email{\{cdolson,ylierler\}@unomaha.edu}
}

\def\titlerunning{Information Extraction Tool \text2alm}
\def\authorrunning{Craig Olson and Yuliya Lierler}

\begin{document}

\maketitle

\newtheorem{lemma}{Lemma}[section]


\label{firstpage}


\begin{abstract}
In this work we design a narrative understanding tool \text2alm. This tool uses an action language~\alm to perform inferences on complex interactions of events described in narratives. The methodology used to implement the \text2alm system was originally outlined by Lierler, Inclezan, and Gelfond~\cite{LierlerIG17} via a manual process of converting a narrative to an \alm model. It relies on a conglomeration of resources and techniques from two distinct fields of artificial intelligence, namely, natural language processing and knowledge representation and reasoning. The effectiveness of system \text2alm  is measured by its ability to correctly answer questions from the \babi tasks published by Facebook Research in 2015. This tool  matched or exceeded the performance of state-of-the-art machine learning methods in six of the seven tested tasks. We also illustrate that the \text2alm approach generalizes to a broader spectrum of narratives.
\end{abstract}


\section{Introduction}
The field of Information Extraction (IE) is concerned with gathering snippets of meaning from text and storing the derived data in structured, machine interpretable form. Consider a sentence
\small
\begin{quote}
{\em  BBDO South in Atlanta, which handles corporate advertising for Georgia-Pacific, will assume additional duties for brands like Angel Soft, said Ken Haldin, a spokesman for Georgia-Pacific from Atlanta.}
\end{quote}
\normalsize
A sample IE system that focuses on identifying organizations and their corporate locations may extract the following predicates from this sentence:
\small
$$
\ba{l}
locatedIn(BBDOSouth,Atlanta)~~~
locatedIn(GeorgiaPacific,Atlanta)\\
\ea
$$
\normalsize
These predicates can then be stored either in a relational database or a logic program, and queried accordingly by well-known methods in computer science. Thus, IE allows us to turn unstructured data present in text into structured data easily accessible for automated querying.

In this paper, we focus on an IE system that is capable of processing simple narratives with {\em action verbs}, in particular,  verbs that express  physical  acts such as \verb{go}, \verb{give}, and \verb{put}. 
Consider a sample narrative that we refer to as the \js discourse:

\small
\begin{align}
\hbox{John traveled to the hallway.}\label{js1} \\
\hbox{Sandra journeyed to the hallway.}\label{js2}
\end{align}
\normalsize

The actions \verb{travel} and \verb{journey} in the narrative describe changes to the narrative's environment, and can be coupled with the reader’s commonsense knowledge to form and alter the reader’s mental picture for the narrative. For example, after reading sentence~\eqref{js1}, a human knows that \textit{John} is the subject of the sentence and \action{traveled} is an action verb describing an action performed by \textit{John}. A human also knows that \action{traveled} describes the act of motion, and specifically that \textit{John}'s location changes from an arbitrary initial location to a new destination, the \textit{hallway}. Lierler et al.~\cite{LierlerIG17} outline a methodology for constructing a Question Answering (QA) system by utilizing IE techniques. Their methodology focuses on performing inferences using the complex interactions of events in narratives. Their process utilizes an action language \alm~\cite{ig16} and an extension of the \verbnet lexicon~\cite{Palmer18,KipperPhd05}. Language \alm enables a system to structure knowledge regarding complex interactions of events and implicit background knowledge in a straight-forward and modularized manner. The knowledge represented in \alm is processed by means of logic programming under answer set semantics and can be used to derive inferences about a given text. The proposed methodology in ~\cite{LierlerIG17} assumes the extension of the \verbnet lexicon with interpretable semantic annotations in \alm. The \verbnet lexicon groups English verbs into classes allowing us to infer that such verbs as \verb{travel} and \verb{journey} practically refer to the same class of events.
 
The processes described in~\cite{LierlerIG17} are exemplified via two sample narratives processed manually. The authors translated those narratives to \alm programs by hand and wrote the supporting \alm modules to capture knowledge as needed. To produce \alm system descriptions for considered narratives, the method by Lierler et al.~\cite{LierlerIG17} utilizes NLP resources, such as semantic role labeler \lth~\cite{lth}, parser and co-reference resolution tools of \core~\cite{ManningSBFBM14}, and lexical resources \propbank~\cite{propbank} and \semlink~\cite{BonialSP13}. Ling~\cite{Ling18} used these resources to automate parts of the method in the \textdrs system. In particular, \textdrs extracts entities, events, and their relations from a given action-based narrative. A narrative understanding system developed within this work, \text2alm, utilizes \textdrs and automates the remainder of the method outlined in~\cite{LierlerIG17}. When considering the \js discourse as an example, system \text2alm produces a set of facts in spirit of the following:
\vspace{-5mm}

\small
\begin{align}
move(john,hallway, 0)~~ move(sandra,hallway, 1)\label{mjs1}\\
loc\_in(john, hallway, 1)~~ loc\_in(john, hallway, 2)~~ loc\_in(sandra, hallway, 2),\label{mjs2}
\end{align} 
\normalsize

where $0, 1, 2$ are time points associated with occurrences of described actions in the \js discourse. Intuitively, time point 0 corresponds to a time prior to utterance of sentence~\eqref{js1}. Time point 1 corresponds to a time upon the completion of the event described in~\eqref{js1}. Facts in~\eqref{mjs1} and~\eqref{mjs2} allow us to provide grounds for answering  questions related to the \js discourse such as:\\

\vspace{-3mm}
\hspace{-5mm}
\small
\begin{tabular}{ll}
	Question:& Ground:\\
	\hline
	Is {\em John} inside the {\em hallway} at the end of the story (time  2)?&$loc\_in(john,hallway,2)$\\
	\hline
	Who is in the {\em hallway} at the end of the story?&$loc\_in(John,hallway,2)$\\
	&$loc\_in(sandra,hallway,2)$\\
\end{tabular}
\normalsize

\vspace{1mm}

We note that modern NLP tools and resources prove to be sufficient to extract facts~\eqref{mjs1} given the \js discourse. Yet, inferring facts such as~\eqref{mjs2} requires complex reasoning about specific actions present in a given discourse and modeling such common sense knowledge as {\em inertia axiom} (stating that {\em things normally stay as they are})~\cite{Leibniz95}. System \text2alm combines the advances in NLP and  knowledge representation and reasoning (KRR) to tackle the complexities of converting  narratives such as the \js discourse into a structured form such as facts in~(\ref{mjs1}-\ref{mjs2}).

The effectiveness of system \text2alm is measured by its ability to answer questions from the \babi tasks~\cite{WestonBCM15}. These tasks were proposed by Facebook Research in 2015 as a benchmark for evaluating basic capabilities of QA systems in twenty categories. Each of the twenty \babi QA tasks is composed of narratives and questions, where 1000 questions are given in training set and 1000 questions are given in a testing set. We extend the information extraction component of the \text2alm by a specialized QA processing module to tackle seven of the \babi tasks containing narratives with action verbs. Tool \text2alm  matched or exceeded the performance of modern machine learning methods in six of these tasks. We also illustrate that the \text2alm approach generalizes to a broader spectrum of narratives than present in \babi.

We start the paper by a review of relevant tools and resources stemming from NLP and KRR communities. We then proceed to describe the architecture of the \text2alm system implemented in this work. We conclude by providing the evaluation data on the system.

\section{Background}
\myparagraph{NLP Resource \verbnet}
\verbnet is a domain-independent English verb lexicon organized into a hierarchical set of verb classes~\cite{Palmer18,KipperPhd05}. The verb classes aim to achieve syntactic and semantic coherence between members of a class. Each class is characterized by a set of verbs and their thematic roles. For example, the verb \verb{run} is a member of the \verbnet class \class{run-51.3.2}. This class is characterized by

\small
\begin{itemize}
\item  96 members  including verbs such as {\em bolt, frolic, scamper,} and {\em weave},
\item four thematic roles, namely, \role{theme}, \role{initial location}, \role{trajectory} and \role{destination},
\item  two subbranches: \class{run-51.3.2-1} and \class{run-51.3.2-2}. For instance, \class{run-51.3.2-2} has members {\em gallop, skip,} and {\em strut}, and has additional thematic roles \role{agent}, \role{result}, and \role{source}.
\end{itemize}
\normalsize

\myparagraph{Dynamic Domains, Transition Diagrams, and Action Language \alm}
{\em Action languages} are formal KRR languages that provide convenient syntactic constructs to represent knowledge about dynamic domains. The knowledge is compiled into a transition diagram, where nodes correspond to possible states of a considered dynamic domain and edges correspond to actions/events whose occurrence signal transitions in the dynamic system. The \js discourse exemplifies a narrative modeling a dynamic domain with three entities \textit{John, Sandra, hallway} and four actions, specifically:

\small
\begin{tabular}{ll}
	$\ajin$ --\textit{ John} travels into the  \textit{hallway}&$\ajout$ --\textit{ John} travels out of the  \textit{hallway}\\
	$\asin$ --\textit{ Sandra} travels into the  \textit{hallway}&$\asout$ --\textit{ Sandra} travels out of the  \textit{hallway}\\
\end{tabular}
\normalsize

Scenarios of a dynamic domain correspond to {\em trajectories} in the domain's transition diagram. Trajectories are sequences of alternating states and actions. A trajectory captures the sequence of events, starting with the initial state associated with time point 0. Each edge is associated with the time point incrementing by 1. 

In this work we utilize an advanced action language \alm~\cite{ig16} to model dynamic domains of given narratives. This language can represent knowledge pertaining to the commonalities of similar actions through means of logic programming under answer set semantics. This is a crucial feature of the language that made it especially fit for this work. In addition, there are efficient solving techniques available for \alm.  In particular, a translation from \alm to logic programs under answer set semantics (answer set programs) was proposed by Inclezan and Gelfond in~\cite{ig16}. In turn, answer set programming is a prominent subfield of automated reasoning supported by a plead of efficient answer set solvers (tools that find solutions to answer set programs). Here we use system \calm~\cite{WertzCZ18} that translates \alm theories into answer set programs in the language of answer set solver {\sc sparc}~\cite{bal13}. We then use system {\sc sparc} to find solutions for \alm theories of interest. 

We illustrate the syntax and semantics of \alm using the \js discourse dynamic domain by first defining an \alm~{\em system description} and then an \alm~{\em history} for this discourse. In language \alm, a dynamic domain is described via a {\em system description} that captures a transition diagram specifying the behavior of a given domain. An \alm system description consists of a theory and a structure. A {\em theory} is comprised of a hierarchy of modules, where a module represents a unit of general knowledge describing relevant sorts, properties, and the effects of actions. The {\em structure} declares instances of entities and actions of the domain. Figure~\ref{fig:jsalm} illustrates these concepts with the \alm formalization of the \js discourse domain.

\begin{figure}
	\footnotesize
	\begin{verbatim}
system description JS_discourse
  theory JS_discourse_theory
    module JS_discourse_module
      sort declarations
        points, agents :: universe	
        move :: actions
          attributes
            actor : agents -> booleans
            origin : points	-> booleans
            destination : points -> booleans
      function declarations
        fluents
          basic
            loc_in : agents * points -> booleans
      axioms
        dynamic causal laws
          occurs(X) causes loc_in(A,D) if instance(X,move), actor(X,A),
                                  destination(X,D).
        executability conditions
          impossible occurs(X) if instance(X,move), actor(X,A), loc_in(A,P),
                                  origin(X,O), P!=O.
          impossible occurs(X) if instance(X,move), actor(X,A), loc_in(A,P),
                                  destination(X,D), P=D.
  structure john_and_sandra
    instances
      john, sandra in agents
      hallway in points
      ajin in move
        actor(john) = true
        destination(hallway) = true
      asin in move
        actor(sandra) = true
        destination(hallway) = true
	\end{verbatim}
	\caption{An \alm system description formalizing the \js discourse dynamic domain\label{fig:jsalm}}
	\normalsize
\end{figure}

The \js discourse theory uses a single module to represent the knowledge relevant to the domain. The module declares the sorts (\sort{agents}, \sort{points}, \sort{move}) and the property (\fluent{loc\_in}) to represent entities and attributes of the domain. \sort{Actions} utilize attributes to define the roles of participating entities. For instance, \attribute{destination} is an attribute of \action{move} that denotes the final location of the mover. Here we ask a reader to draw a parallel between the notions of an attribute and a \verbnet thematic role.

The \js discourse theory also defines two types of axioms, dynamic causal laws and executability conditions, to represent commonsense knowledge associated with a \action{move} action. The dynamic causal law states that if a \action{move} action occurs with a given \attribute{actor} and \attribute{destination}, then the \attribute{actor}'s location becomes that of the \attribute{destination}. The executability conditions restrict an action from occurring if the action is an instance of \action{move}, where the \attribute{actor} and \attribute{actor}'s location are defined, but either (i) the \attribute{actor}'s location is not equal to the \attribute{origin} of the \action{move} event or (ii) the \attribute{actor}'s location is already the \attribute{destination}.

An \alm structure in Figure~\ref{fig:jsalm} defines the entities and actions from the \js discourse. For example, it states that \textit{john} and \textit{sandra} are \sort{agents}. Also, action $\ajin$ is declared as an instance of \action{move} where \textit{john} is the \attribute{actor} and \textit{hallway} is the \attribute{destination}.

An \alm system description can be coupled with a {\em history}. A history is a particular scenario described by observations about the values of properties and occurring events. In the case of narratives, a history describes the sequence of events by stating occurrences of specific actions at given time points. For instance, the \js discourse history contains the events
\begin{itemize}
	\item \textit{John} \textit{moves} to the \textit{hallway} at the beginning of the story (an action $\ajin$ occurs at time 0)  and 
	\item \textit{Sandra} \textit{moves} to the \textit{hallway} at the next point of the story (an action $\asin$ occurs at time~1).	
\end{itemize}
\vspace{1mm}
The following history is appended to the end of the system description in Figure~\ref{fig:jsalm} to form an \alm program for the \js disocurse. We note that $happened$ is a keyword  that captures the occurrence of actions.

\vspace{-1mm}
\small
\begin{verbatim}
  history
    	happened(ajin, 0).
    	happened(asin, 1).
\end{verbatim}
\vspace{-1mm}
\normalsize

\myparagraph{An \alm Solver \calm}
System \calm is an \alm solver developed at Texas Tech University by Wertz, Chandrasekan, and Zhang~\cite{WertzCZ18}. It uses an \alm program to produce a ''model'' for an encoded dynamic domain. The engine for system \calm (i) constructs a logic program under stable model/answer set semantics~\cite{gel88}, whose answer sets/solutions are in one-to-one correspondence with the models of the \alm program, and (ii) uses an answer set solver {\sc sparc}~\cite{bal13} for finding these models. In this manner, \calm processes the knowledge represented by an \alm program to enable reasoning capabilities. The \alm program in Figure~\ref{fig:jsalm} follows the \calm syntax. However, system \calm requires two additional components for this program to be executable. The user must specify (i) the computational task and (ii) the max time point considered. 

In our work we utilize the fact that system \calm can solve a task of temporal projection, which is the process of determining the effects of a given sequence of actions executed from a given initial situation (which may be not fully determined). In the case of a narrative the initial situation is often unknown, whereas the sequence of actions are provided by the discourse. Inferring the effects of actions allows us to answer questions about the narrative's domain. We insert the following statement in the \alm program prior to the history to perform temporal projection:

\small
\begin{verbatim}
  temporal projection
\end{verbatim}
\normalsize
Additionally, \calm requires the max number of steps to be stated. Intuitively, we see this number as an upper bound on the ''length'' of considered trajectories. This information denotes the final state's time point in temporal projection problems. We insert the following line in the \alm program to define the max steps for the \js discourse \alm program:

\small
\begin{verbatim}
  max steps 3
\end{verbatim}
\normalsize

For the case of the temporal projection task, a model of an \alm program is a trajectory in the transition system captured by the \alm program that is ''compatible'' with the provided history. A compatible model correlate to the answer set solved by \calm.
For the \js discourse \alm program, the \calm computes a model that includes the following expressions:

\small
\begin{verbatim}
  happened(ajin, 0),        happened(asin, 1),	
  loc_in(john, hallway, 1), loc_in(sandra, hallway, 2), loc_in(john, hallway, 2)
\end{verbatim}
\normalsize
	
\myparagraph{\alm Knowledge Base \corealmlib}
The \corealmlib is an \alm library of generic commonsense knowledge for modeling dynamic domains developed by Inclezan~\cite{Inclezan16}. The library's foundation is the Component Library or CLib~\cite{BarkerPC01}, which is a collection of general, reusable, and interrelated components of knowledge. CLib was populated with knowledge stemming from linguistic and ontological resources, such as \verbnet, \wordnet, \framenet, a thesaurus, and an English dictionary. The \corealmlib was formed by translating CLib into \alm to obtain descriptions of 123 action classes grouped into 43 reusable modules. The modules are organized into a hierarchical structure, and contain action classes and axioms to support commonsense reasoning. 
An example of one such axiom from the \textit{motion} module is provided in Figure~\ref{fig:motionaxiom}. This axiom states that if a \textit{move} action occurs where \textit{O} is the object moving and \textit{D} is a \textit{spatial entity} and the destination, then the \textit{location} of \textit{O} becomes \textit{D}.

\begin{figure}
\footnotesize
\begin{verbatim}	
  occurs(X) causes location(O, D) if instance(X, move),
                                                 object(X, O),
                                                 destination(X, D),
                                                 instance(D, spatial_entity).
\end{verbatim}
\caption{Commonsense Knowledge Axiom from the \corealmlib \textit{motion} module.\label{fig:motionaxiom}}
\normalsize
\end{figure}

\section{System \text2alm Architecture}
\begin{figure}
  \centering
  \includegraphics[scale=0.55]{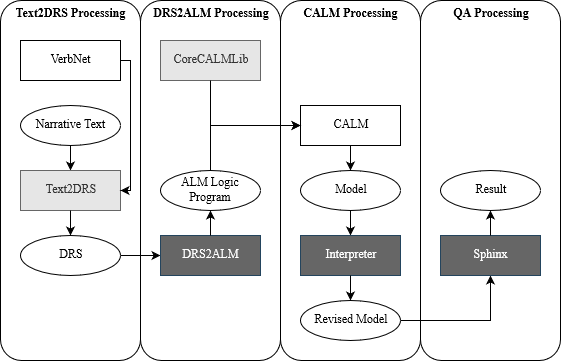}
	\caption{System \text2alm Architecture\label{fig:arch}}
\end{figure}

Lierler, Inclezan, and Gelfond~\cite{LierlerIG17} outline a methodology for designing IE/QA systems to make inferences based on complex interactions of events in narratives. This methodology is exemplified with two sample narratives completed manually by the authors. System \text2alm automates this process. Figure~\ref{fig:arch} pretenses the architecture of the system. It implements four main tasks/processes: 

\small
\begin{enumerate}
	\item \textdrs Processing -- Entity, Event, and Relation Extraction
	\item \drsalm Processing -- Creation of \alm Program
	\item \calm Processing -- \alm Model Generation and Interpretation
	\item QA Processing
\end{enumerate}
\normalsize

Figure~\ref{fig:arch} denotes each process by its own column. Ovals identify inputs and outputs. Systems or resources are represented with white, grey, and black rectangles. White rectangles denote existing, unmodified resources. Grey rectangles are used for existing, but modified resources. Black rectangles signify newly developed subsystems. The first three processes form the core of \text2alm, seen as an IE system. The QA Processing component is specific to the \babi QA benchmark that we use to illustrate the validity of the approach advocated by \text2alm. The system's source code is available at \url{https://github.com/cdolson19/Text2ALM}.

\small
\normalsize

\subsection{\textdrs Processing}
The method by Lierler et al.~\cite{LierlerIG17} utilizes NLP resources, such as semantic role labeler \lth~\cite{lth}, parsing and coreference resolution tools of \core~\cite{ManningSBFBM14}, and lexical resources \propbank~\cite{propbank} and \semlink~\cite{BonialSP13} to produce \alm system descriptions for considered narratives. System \textdrs~\cite{Ling18} was developed with these resources to deliver a tool that extracts entities, events, and their relations from given narratives. The \textdrs tool formed the starting point in the development of \text2alm due to its ability to extract basic entity and relational information from a narrative. The output of the \textdrs system is called a discourse representation structure, or DRS~\cite{kampreyle93}. A DRS captures key information present in discourse in a structured form. For example, Figure~\ref{fig:drs} presents the DRS for the \js discourse. 

\begin{figure}
	\small
	\begin{verbatim}	
entity(r1).           entity(r2).               entity(r3). 	
property(r1, "John"). property(r2, "hallway"). 	property(r3, "Sandra"). 
	
event(e1). 	                           event(e2). 	
eventType(e1, "run-51.3.2-1").         eventType(e2, "run-51.3.2-1"). 	
eventTime(e1, 0).                      eventTime(e2, 1). 	
eventArgument(e1, "Theme", r1).        eventArgument(e2, "Theme", r3).
eventArgument(e1, "Destination", r2).  eventArgument(e2, "Destination", r2).
	\end{verbatim}
	\caption{DRS for the \js discourse.\label{fig:drs}}
	\normalsize
\end{figure}

This DRS states that there are three entities and two events that take part in the \js narrative. The DRS assigns names, or referents, to the entities ($r1,r2,$ and $r3$) and the events ($e1, e2$). For instance, entity $r1$ and event $e1$ denote \textit{John} and an event representing the \verbnet class \class{run-51.3.2-1}, respectively. The \role{theme} (which is one of the thematic roles associated with \class{run-51.3.2-1}) of event $e1$ is entity $r1$ (or, \textit{John}) and the \role{destination} of this event is entity~$r2$ (or, \textit{hallway}). Event $e1$ occurs at time point 0, while event $e2$ occurs at time point 1. We refer an interested reader to the work by Ling~\cite{Ling18} for the details of the \textdrs component. In realms of this project, \textdrs was modified to accommodate \verbnet v 3.3 (in place of \verbnet v 2), which provides broader coverage of verbs.

\subsection{\drsalm Processing}\label{sec:drsalmproc}
The \drsalm subsystem is concerned with combining commonsense knowledge related to events in a discourse with the information from the DRS generated by \textdrs. The goal of this process is to produce an \alm program consisting of a system description and a history for the scenario described by the narrative. The system description is composed of a theory containing relevant commonsense knowledge and a structure that is unique for a given narrative. Since the structure is specific to a given narrative, it is created using the information from a narrative's DRS. Meanwhile the theory represents the commonsense knowledge associated with a narrative's actions. Thus, the theory depends on a general, reusable knowledge base pertaining to actions. The \corealmlib knowledge base was modified to form \corecalmlib to fit this need of the \text2alm system. We organize this section by (1) explaining how \corecalmlib was obtained and (2) provide details on how a narrative's \alm program is generated.

\myparagraph{Library \corecalmlib} 
To obtain the \corecalmlib knowledge base, the following modifications to the \corealmlib were made:

\begin{tabular}{lcl}
	1. Syntactic adjustments & & 3. \verbnet extensions\\
	2.  Property extractions & & 4.  Axiom changes\\
\end{tabular}

First, syntactic adjustments were implemented to make the library compatible with the \calm syntax. Second, we observed that the \corealmlib has instances where properties (fluents) with the same name are declared in multiple modules. Yet semantically, these properties are assumed to be the same across all modules. We found this approach counter-intuitive from the point of knowledge-base design, thus we extracted all fluent declarations from \corealmlib modules and created new modules whose purpose was to declare fluents. These modules were organized by their properties and grouped similar properties together. The  original \corealmlib modules now import the necessary properties as needed. Regarding \verbnet extensions, \corealmlib was further modified by adding a module for every \verbnet class we observed in the \babi QA task training sets. We discuss these training sets in detail in Section~\ref{sec:evaluation}. In particular, 52 of \verbnet's 274 classes were formalized with modules in \corecalmlib. Each \verbnet module defines a sort for that verb class that inherits from one of the 123 action classes stemming from \corealmlib. Specifically, we utilize 15 action classes formalized in \corealmlib, stemming from 9 of its total 43 modules. Thematic roles from the \verbnet lexicon are then mapped via state constraints to the attributes associated with actions already used by the \corealmlib library. These \verbnet modules are stored in a \corecalmlib sub-library that we call \vnclasslib. Lastly, we modified and added axioms into some \corealmlib modules after identifying pieces of knowledge that were not represented within the original library. When not considering fluent extractions, fluents were altered or added to only four modules from the original \corealmlib. This supports the hypothesis that \corealmlib can provide an effective baseline for commonsense reasoning about actions. All modifications to the \corealmlib to form \corecalmlib are explained further in~\cite{Olson19}.

\myparagraph{\alm Program Generation}
The \drsalm processing step generates an \alm program for a given discourse by combining the information in a narrative's DRS and the \corecalmlib library. We first examine the theory in the program's system description. We start by identifying the general knowledge associated with a narrative's domain by importing the \verbnet modules from the \corecalmlib for all \verbnet classes associated with a narrative. These provide the commonsense knowledge backbone for the actions in the narrative. Then, we define a new module unique to the narrative. This module declares entities from the narrative as new sorts inheriting from base \corecalmlib sorts. We chose to declare the narrative's entities as new sorts to provide more flexibility to define additional, unique attributes associated with the entities if the need arises. However, to declare these new sorts we must identify the \corecalmlib parent sort to inherit from. We rely on the \verbnet thematic roles associated with an entity to make this selection. We grouped \verbnet thematic roles into four parent sorts of \corecalmlib by reviewing the thematic roles associated with the \verbnet classes in the training sets and attempting to map these to the most similar sorts defined by the original \corealmlib.  Figure~\ref{fig:RolesToSorts} presents the groupings. If an entity is associated with roles from different categories, we use a prioritized sort order defined as follows:

\vspace{-1mm}

\begin{center}$living\_entity >> place >> spatial\_entity >> entity,$
	\end{center}	

\vspace{-1mm}
\noindent
where $>>$ is transitive and states that the left argument has a higher priority than the right one.

\begin{figure}
  \centering
	\small
	\begin{tabular}{l|l|l|l|l}
		Parent Sort & living\_entity & place & spatial\_entity & entity\\
		\hline
		Associated & Actor, Agent & Location & Destination & Instrument, Material \\
		Thematic & Beneficiary & Place & Initity\_location & Pivot, Product Duration\\
		Roles & Cause, Co-Agent & & Source & Stimulus, Time, Extent \\
		& Co-Theme, Recipient& & & Trajectory, Initial\_time\\
		& Experiencer & & & Topic, Value,  Goal\\
		& Participant, Patient & & & Result, Attribute \\
		& Theme, Undergoer & & & Final\_time, Frequency \\
	\end{tabular}
	\caption {\verbnet Thematic Roles to \corealmlib Sorts \label{fig:RolesToSorts}}
	\normalsize
\end{figure}

We now turn our attention to the process of generating the structure and history for the \alm program. The structure declares the specific entities and events from the narrative. Entity IDs from a given narrative's DRS are defined as instances of the corresponding entity sort from the theory. Events are also declared as instances of their associated \verbnet class sort, and the entities related to events are listed as attributes of these events. The history states the order and timepoints in which narrative's events happened. We extract this information from the arguments expressed in DRS. To exemplify the described process, Figure~\ref{fig:jsalmauto} presents the \alm program output by the \drsalm Processing stage applied towards the \js discourse DRS in Figure~\ref{fig:drs}. Note that the \alm theory in Figure~\ref{fig:jsalmauto} imports the \verbnet module for \class{run-51.3.2-1} from the \vnclasslib. The two events in the \js discourse were identified as members of the \verbnet class \class{run-51.3.2-1}. Thus, the module associated with this class is imported to retrieve the knowledge relevant to \class{run} events in the \js discourse domain. 

\begin{figure}
	\footnotesize
	\begin{verbatim}
system description js_discourse
  theory js_discourse_theory
    import t_run_51_32.m_run_51_3_2_1 from VN_class_library
    module js_discourse
      depends on t_run_51_3_2.m_run_51_3_2_1
      sorts declarations
        john :: living_entity
        hallway :: spatial_entity
        sandra :: living_entity
  structure js_discourse_structure
    instances
      r1 in john
      r2 in hallway
      r3 in sandra
      e1 in run_51_3_2_1
        vn_theme(r1) = true
        vn_destination(r2) = true
      e2 in run_51_3_2_1
        vn_theme(r3) = true
        vn_destination(r2) = true
temporal projection
max steps 3
history
  happened(e1,0).
  happened(e2 1).
	\end{verbatim}
	\caption{An \alm system description automatically created by the \drsalm processing\label{fig:jsalmauto}}
	\normalsize
\end{figure}

\subsection{\calm and QA Processing}
In the \calm Processing performed by \text2alm, the \calm system is invoked on a given narrative's \alm program that was generated by the \drsalm Processing stage. The \calm system computes a model via logic programming under answer set semantics. We then perform post-processing on this model to make its content more readable for a human by replacing all entities IDs with their names from the narrative. For instance, given the \alm program in Figure~\ref{fig:jsalmauto}, the output of the \calm Processing will include expressions: 

{\tt loc\_in(John,hallway,1), loc\_in(John,hallway,2), loc\_in(Sandra,hallway,2).}\\
We note that no other {\tt loc\_in} fluents will be present in the output.

A model derived by the \calm system contains facts about the entities and events from the narrative supplemented with basic commonsense knowledge associated with the events. We use a subset of the \babi QA tasks to test the \text2alm system's IE effectiveness and implement QA capabilities within the \sphinx subsystem (see Figure~\ref{fig:arch}). It utilizes regular expressions to identify the kind of question that is being asked and then query the model for relevant information to derive an answer. The \sphinx system is specific to the \babi QA task and is not a general purpose question answering component.

Additional information on the components of system \text2alm  are given in~\cite{Olson19}.

\section{\text2alm Evaluation}\label{sec:evaluation}
\paragraph{Related Work:}
Many modern QA systems predominately rely on machine learning techniques. However, there has recently been more work related to the design of QA systems combining advances of NLP and KRR. The \text2alm system is a representative of the latter approach. Other approaches include the work by Clark, Dalvi, and Tandon~\cite{ClarkDT14} and Mitra and Baral~\cite{MitraB16}. Mitra and Baral~\cite{MitraB16} use a training dataset to learn the knowledge relevant to the action verbs mentioned in the dataset. They posted nearly perfect test results on the \babi tasks. However, this approach doesn't scale to narratives that utilize other action verbs which are not present in the training set, including synonymous verbs. For example, if their system is trained on \babi training data that contains verb \verb{travel}it will process the \js discourse correctly. Yet, if we alter the \js discourse by exchanging \verb{travel} with a synonymous word \verb{stroll}, their system will fail to perform inferences on this altered narrative (note that \verb{stroll} does not occur in the \babi training set). We address this limitation in the \text2alm system because the system does not rely upon the training narratives for the commonsense knowledge. If the verbs occurring in narratives belong to  \verbnet classes whose semantics have been captured within \corecalmlib then \text2alm is normally able to process them properly.

Another relevant QA approach is the work by Clark, Dalvi, and Tandon~\cite{ClarkDT14}. This approach uses \verbnet to build a knowledge base containing rules of preconditions and effects of actions utilizing the semantic annotations that \verbnet provides for its classes. In our work, we can view \alm modules associated with \verbnet classes as machine interpretable alternatives to these annotations. However, Clark et al.~\cite{ClarkDT14} use the first and most basic action language \strips~\cite{FikesN71} for inference. The \strips language allows more limited capabilities than the \alm language in modeling complex interactions between events.

\myparagraph{Evaluation}
We use a subset of Facebook AI Research’s \babi dataset~\cite{WestonBCM15} to evaluate system \text2alm. These tasks were proposed by Facebook Research in 2015 as a benchmark for evaluating basic capabilities of QA systems in twenty categories. Each of the twenty \babi QA tasks is composed of narratives and questions, where 1000 questions are given in training set and 1000 questions are given in a testing set. The goal of the tasks are to answer the questions in the testing sets correctly while also minimizing the number of questions used from the training set to develop a solution. We evaluate  the \text2alm system with all 1000 questions in the testing sets for tasks 1, 2, 3, 5, 6, 7, and 8. These tasks are selected for two reasons. First, these tasks contain action-based narratives that are of focus in this work. Second, the underlying IE engine \textdrs requires further development to formulate representations of more advanced sentence structures, such as those containing negation and indefinite knowledge. Figure~\ref{fig:babi} provides an example of a narrative and a question from the training set of \babi task 2-Two Supporting Facts. For this task, a QA system must combine information from two sentences in the given narrative. The narrative in Figure~\ref{fig:babi} consists of six sentences. A question is given in line 7, followed by the answer and identifiers for the two sentences that provide information to answer the question.

\footnotesize
\begin{figure}
	\texttt{
\begin{tabular}{ll}
	1 Mary moved to the bathroom.&	5 Mary went back to the kitchen.\\
	2 Sandra journeyed to the bedroom.& 6 Mary went back to the garden.\\
	3 Mary got the football there. &7 Where is the football? garden 3 6\\
	4 John went to the kitchen.&
\end{tabular}
}
	\caption {Example narrative and question from \babi task 2 training set
		\label{fig:babi}}
\end{figure}
\normalsize

The \babi dataset enables us to compare \text2alm's IE/QA ability with other modern approaches designed for this task. The left hand side of Figure~\ref{fig:evaluation} compares the accuracy of the \text2alm system with the machine learning approach AM+NG+NL MemNN described by~Weston~et al.~\cite{WestonBCM15}. In that work, the authors  compared results from 8 machine learning approaches on \babi tasks and the AM+NG+NL MemNN (Memory Network) method performed best almost across the board. There were two exceptions among the seven tasks that we consider. For the Task 7-Counting the  AM+N-GRAMS MemNN  algorithm was reported to obtain a higher accuracy of 86\%. Similarly, for the Task 8-Lists/Sets the  AM+NONLINEAR MemNN  algorithm was reported to obtain accuracy of 94\%. Figure~\ref{fig:evaluation} also presents the details on the Inductive Rule Learning and Reasoning (IRLR) approach by~\cite{MitraB16}. We cannot compare \text2alm performance with the methodology by~\cite{ClarkDT14} because their system is not available and it has not been evaluated using the \babi tasks.

\small
\begin{figure}
  \centering
	\begin{tabular}{l|c|c|c}
				&\multicolumn{3}{c}{Accuracy}\\
				\hline			
		\babi Task & AM+NG+NL Mem NN & IRLR & {\text2alm} \\
		\hline
		1-Single Sup. Facts&100&100&100\\
		\hline
		2-Two Sup. Facts&100&100&100\\
		\hline
		3-Three Sup. Facts&100&100&100\\
		\hline
		5-Three Arg. Rels.&98&100&22\\
		\hline
		6-Yes/No&100&100&100\\	
		\hline
		7-Counting&85&100&96.1\\
		\hline
		8-Lists/Sets&91&100&100\\
	\end{tabular}
	\caption{System Evaluations \label{fig:evaluation}}
\end{figure}
\normalsize

System \text2alm matches the Memory Network approach by Weston et al.~\cite{WestonBCM15} at 100\% accuracy in tasks 1, 2, 3, and 6 and performs better on tasks 7 and 8. When compared to the methodology by Mitra and Baral~\cite{MitraB16}, the Text2ALM system matches the results for tasks 1, 2, 3, 6, and 8, but is outperformed in tasks~5 and 7. 

The results of the \text2alm system were comparable to the industry-leading results with one outlier, namely, task 5. We investigated the reason. It turns out that the testing set frequently contained a phrase of the form:

\small
\begin{tabular}{l r}
	\hbox{{\tt Entity1} \textit{handed the} {\tt Object} \textit{to} \tt{Entity2}.}~~~~~~&\hbox{e.g., {\it Fred handed the football to Bill.}}\\
\end{tabular}
\normalsize

The \text2alm system failed to properly process such phrases  because the semantic role labeler \lth, a subcomponent of the \textdrs system, incorrectly annotated the sentence. In particular, \lth consistently considered a reading in spirit of the following: {\it Fred handed Bill's football away}. This annotation error prevents  \textdrs  from adding crucial event argument to the DRS stating that {\tt Entity2} plays the thematic role of \role{destination} in the phrase. Consequently, the \text2alm system does not realize that possession of the object was passed from {\tt Entity1} to {\tt Entity2}.

Even though our system does not match the scores and breadth of testing as the approach by Mitra and Baral~\cite{MitraB16}, who tested on all 20 \babi tasks, we consider the scores obtained by \text2alm interesting for several reasons. First, the approach implemented by \text2alm suggests that lexical resources, such as \verbnet, \propbank, and \semlink can be utilized effectively to support IE and KRR tasks. Second, the scores support the hypothesis that a relatively few, but general, commonsense rules about the effects of actions can be used to generalize to a broad number of similar actions. To illustrate that the approach by system \text2alm generalizes well, we create a variant of the \babi task, which we call \babiplus. We obtain \babiplus by changing $50\%$ of action verbs occurring in testing set narratives with their synonymous counterparts. For example, $50\%$ of instances of \verb{travelled} and \verb{grabbed} were replaced with \verb{sprinted} and \verb{seized}, respectively. In total, 13 synonymous verbs were introduced. To ensure that \text2alm system handles the \babiplus tasks, two extensions were made to its resources. First, the \corecalmlib knowledge base was augmented with appropriate mappings for two more \verbnet classes. Second, two new entries in \semlink were introduced for verbs lacking the mappings and four modifications to existing \semlink entries were required (where \semlink is a key resource of the \textdrs system). Figure~\ref{fig:plus_evaluation} presents the accuracy of the \text2alm system on the \babiplus tasks. 

\small
\begin{figure}
  \centering
	\begin{tabular}{l|l||l|l}		
		\babiplus Task &  Accuracy & \babiplus Task &  Accuracy\\
		\hline
		1-Single Sup. Facts&97.9
				&6-Yes/No&99\\	
					
		\hline
		2-Two Sup. Facts&97.8&
		7-Counting&96.1\\
		\hline
		3-Three Sup. Facts&97.4&
				
				8-Lists/Sets&100\\
				
		\hline
		5-Three Arg. Rels.&19&\multicolumn{2}{l}{}\\
	\end{tabular}
	
	\caption{{\text2alm} Evaluation on \babiplus Tasks \label{fig:plus_evaluation}}
\end{figure}
\normalsize

\section{Conclusion and Future Work}
Lierler, Inclezan, and Gelfond~\cite{LierlerIG17} outline a methodology for designing IE/QA systems to make inferences based on complex interactions of events in narratives. To explore the feasibility of this methodology, we built the \text2alm system to take an action-based narrative as input and output a model encoding facts about the given narrative. 
We tested the system over tasks 1, 2, 3, 5, 6, 7, and 8 from the \babi QA dataset~\cite{WestonBCM15}. System \text2alm matched or outperformed the results of modern machine learning methods in all of these tasks except task 5. It also matched the results of another KRR approach~\cite{MitraB16} in tasks 1, 2, 3, 6, and 8, but did not perform as well in tasks 5 and 7. However, our approach adjusts well to narratives with a more diverse lexicon. Additionally, the ability of the \corecalmlib to represent the interactions of events in the \babi narratives serves as a proof of usefulness of the original \corealmlib endeavor. 

We conclude our work by listing future research directions in some areas, (i) Expanding narrative processing capabilities, (ii) Expanding QA ability, (iii) Exploring additional reasoning tasks.

The \babi QA tasks provided basic narratives to evaluate the effectiveness of information extraction by system \text2alm. However, these are basic narratives with simple sentence structures. Future work includes expanding the narrative processing capabilities of system \text2alm, as well as reducing the impact of semantic role labeling errors. We need to enhance the \textdrs subsystem's capabilities in order to provide more detailed IE on narratives. Also, so far we provided \alm annotations via the \corecalmlib library for twenty two classes of \verbnet. In the future we intend to cover all \verbnet classes.

Questions in the bAbI QA tasks follow pre-specified formats. Therefore, system \text2alm's QA ability relies on simple regular expression matching. Further research is required on representing generic questions and answers before using the system's IE abilities in other applications. Additionally, our approach should be tested on more advanced QA datasets, such as  \propara~\cite{MishraDHTYC18}. Conducting tests on the \propara dataset would enable us to compare the results of \text2alm to the approach by~\cite{ClarkDT14}.

Finally, we will build on \text2alm's reasoning abilities. For example, the \calm model may sometimes not contain atoms that could be argued as reasonable. For example,  given a narrative  \textit{The monkey is in the tree. The monkey grabs the banana.}, the \calm model will contain fluents stating that the \textit{monkey}'s location is the \textit{tree} at time point 1, the \textit{monkey} is holding the \textit{banana} at time point 2, and the \textit{banana}'s location is the \textit{tree} at time point 2. However, it is also natural to infer that the \textit{banana}'s location is the \textit{tree} when the \textit{monkey} grabs it (time point 1). Yet, that requires reasoning that goes beyond temporal projection.

\bibliographystyle{eptcs}
\bibliography{verbnet_alm}

\begin{thebibliography}{1}
\providecommand{\bibitemdeclare}[2]{}
\providecommand{\surnamestart}{}
\providecommand{\surnameend}{}
\providecommand{\urlprefix}{Available at }
\providecommand{\url}[1]{\texttt{#1}}
\providecommand{\href}[2]{\texttt{#2}}
\providecommand{\urlalt}[2]{\href{#1}{#2}}
\providecommand{\doi}[1]{doi:\urlalt{http://dx.doi.org/#1}{#1}}
\providecommand{\bibinfo}[2]{#2}

\bibitemdeclare{inproceedings}{bal13}
\bibitem{bal13}
\bibinfo{author}{Evgenii \surnamestart Balai\surnameend},
  \bibinfo{author}{Michael \surnamestart Gelfond\surnameend} \&
  \bibinfo{author}{Yuanlin \surnamestart Zhang\surnameend}
  (\bibinfo{year}{2013}): \emph{\bibinfo{title}{Towards Answer Set Programming
  with Sorts}}.
\newblock In \bibinfo{editor}{Pedro \surnamestart Cabalar\surnameend} \&
  \bibinfo{editor}{Tran~Cao \surnamestart Son\surnameend}, editors: {\sl
  \bibinfo{booktitle}{Logic Programming and Nonmonotonic Reasoning}},
  \bibinfo{publisher}{Springer Berlin Heidelberg}, \bibinfo{address}{Berlin,
  Heidelberg}, pp. \bibinfo{pages}{135--147},
  \doi{10.1007/978-3-540-72200-7\_4}.

\bibitemdeclare{article}{BarkerPC01}
\bibitem{BarkerPC01}
\bibinfo{author}{Ken \surnamestart Barker\surnameend}, \bibinfo{author}{Bruce
  \surnamestart Porter\surnameend} \& \bibinfo{author}{Peter \surnamestart
  Clark\surnameend} (\bibinfo{year}{2001}): \emph{\bibinfo{title}{{A Library of
  Generic Concepts for Composing Knowledge Bases}}}.
\newblock {\sl \bibinfo{journal}{Proceedings of the 1st International
  Conference on Knowledge Capture - K-CAP}}, pp. \bibinfo{pages}{14--21},
  \doi{10.1145/500742.500744}.
\newblock \urlprefix\url{http://portal.acm.org/citation.cfm?doid=500737.500744
  http://www.cs.utexas.edu/users/mfkb/papers/kcap01.pdf}.

\bibitemdeclare{misc}{BonialSP13}
\bibitem{BonialSP13}
\bibinfo{author}{Claire \surnamestart Bonial\surnameend},
  \bibinfo{author}{Kevin \surnamestart Stowe\surnameend} \&
  \bibinfo{author}{Martha \surnamestart Palmer\surnameend}
  (\bibinfo{year}{2013}): \emph{\bibinfo{title}{{SemLink}}}.
\newblock \bibinfo{howpublished}{\url{https://verbs.colorado.edu/semlink/}}.

\bibitemdeclare{article}{ClarkDT14}
\bibitem{ClarkDT14}
\bibinfo{author}{Peter \surnamestart Clark\surnameend},
  \bibinfo{author}{Bhavana \surnamestart Dalvi\surnameend} \&
  \bibinfo{author}{Niket \surnamestart Tandon\surnameend}
  (\bibinfo{year}{2018}): \emph{\bibinfo{title}{What Happened? {L}everaging
  {V}erb{N}et to Predict the Effects of Actions in Procedural Text}}.
\newblock {\sl \bibinfo{journal}{CoRR}} \bibinfo{volume}{abs/1804.05435}.
\newblock \urlprefix\url{http://arxiv.org/abs/1804.05435}.

\bibitemdeclare{article}{FikesN71}
\bibitem{FikesN71}
\bibinfo{author}{Richard~E. \surnamestart Fikes\surnameend} \&
  \bibinfo{author}{Nils~J. \surnamestart Nilsson\surnameend}
  (\bibinfo{year}{1971}): \emph{\bibinfo{title}{{Strips: A new approach to the
  application of theorem proving to problem solving}}}.
\newblock {\sl \bibinfo{journal}{Artificial Intelligence}}
  \bibinfo{volume}{2}(\bibinfo{number}{3-4}), pp. \bibinfo{pages}{189--208},
  \doi{10.1016/0004-3702(71)90010-5}.

\bibitemdeclare{inproceedings}{gel88}
\bibitem{gel88}
\bibinfo{author}{Michael \surnamestart Gelfond\surnameend} \&
  \bibinfo{author}{Vladimir \surnamestart Lifschitz\surnameend}
  (\bibinfo{year}{1988}): \emph{\bibinfo{title}{The stable model semantics for
  logic programming}}.
\newblock In \bibinfo{editor}{Robert \surnamestart Kowalski\surnameend} \&
  \bibinfo{editor}{Kenneth \surnamestart Bowen\surnameend}, editors: {\sl
  \bibinfo{booktitle}{Proceedings of International Logic Programming Conference
  and Symposium}}, \bibinfo{publisher}{MIT Press}, pp.
  \bibinfo{pages}{1070--1080}.

\bibitemdeclare{article}{Inclezan16}
\bibitem{Inclezan16}
\bibinfo{author}{Daniela \surnamestart Inclezan\surnameend}
  (\bibinfo{year}{2016}): \emph{\bibinfo{title}{{CoreALMlib: An ALM library
  translated from the Component Library}}}.
\newblock {\sl \bibinfo{journal}{Theory and Practice of Logic Programming}}
  \bibinfo{volume}{16}(\bibinfo{number}{5-6}), pp. \bibinfo{pages}{800--816},
  \doi{10.1017/S1471068416000363}.

\bibitemdeclare{article}{ig16}
\bibitem{ig16}
\bibinfo{author}{Daniela \surnamestart Inclezan\surnameend} \&
  \bibinfo{author}{Michael \surnamestart Gelfond\surnameend}
  (\bibinfo{year}{2016}): \emph{\bibinfo{title}{Modular action language
  {ALM}}}.
\newblock {\sl \bibinfo{journal}{{TPLP}}}
  \bibinfo{volume}{16}(\bibinfo{number}{2}), pp. \bibinfo{pages}{189--235},
  \doi{10.1017/S1471068415000095}.

\bibitemdeclare{inproceedings}{lth}
\bibitem{lth}
\bibinfo{author}{Richard \surnamestart Johansson\surnameend} \&
  \bibinfo{author}{Pierre \surnamestart Nugues\surnameend}
  (\bibinfo{year}{2007}): \emph{\bibinfo{title}{LTH: Semantic Structure
  Extraction using Nonprojective Dependency Trees}}.
\newblock In: {\sl \bibinfo{booktitle}{Proceedings of the Fourth International
  Workshop on Semantic Evaluations (SemEval-2007)}},
  \bibinfo{publisher}{Association for Computational Linguistics},
  \bibinfo{address}{Prague, Czech Republic}, pp. \bibinfo{pages}{227--230},
  \doi{10.3115/1621474.1621522}.
\newblock \urlprefix\url{http://www.aclweb.org/anthology/S/S07/S07-1048}.

\bibitemdeclare{book}{kampreyle93}
\bibitem{kampreyle93}
\bibinfo{author}{Hans \surnamestart Kamp\surnameend} \& \bibinfo{author}{Uwe
  \surnamestart Reyle\surnameend} (\bibinfo{year}{1993}):
  \emph{\bibinfo{title}{From discourse to logic}}.
\newblock \bibinfo{volume}{1,2}, \bibinfo{publisher}{Kluwer},
  \doi{10.1007/978-94-011-2066-1}.

\bibitemdeclare{phdthesis}{KipperPhd05}
\bibitem{KipperPhd05}
\bibinfo{author}{Karin \surnamestart Kipper-Schuler\surnameend}
  (\bibinfo{year}{2005}): \emph{\bibinfo{title}{Verb{N}et: A Broad-Coverage,
  Comprehensive Verb Lexicon}}.
\newblock Ph.D. thesis, \bibinfo{school}{University of Pennsylvania}.

\bibitemdeclare{article}{Leibniz95}
\bibitem{Leibniz95}
\bibinfo{author}{Gottfried~Wilhelm \surnamestart Leibniz\surnameend}
  (\bibinfo{year}{1995}): \emph{\bibinfo{title}{{Philosophical Writings}}}.
\newblock {\sl \bibinfo{journal}{Everyman}}.

\bibitemdeclare{inproceedings}{LierlerIG17}
\bibitem{LierlerIG17}
\bibinfo{author}{Yuliya \surnamestart Lierler\surnameend},
  \bibinfo{author}{Daniela \surnamestart Inclezan\surnameend} \&
  \bibinfo{author}{Michael \surnamestart Gelfond\surnameend}
  (\bibinfo{year}{2017}): \emph{\bibinfo{title}{Action Languages and Question
  Answering}}.
\newblock In: {\sl \bibinfo{booktitle}{{IWCS} 2017 - 12th International
  Conference on Computational Semantics - Short papers}}.

\bibitemdeclare{misc}{Ling18}
\bibitem{Ling18}
\bibinfo{author}{Gang \surnamestart Ling\surnameend} (\bibinfo{year}{2018}):
  \emph{\bibinfo{title}{{From Narrative Text to VerbNet-Based DRSes: System
  Text2DRS}}}.
\newblock \bibinfo{howpublished}{Project Report,
  \url{https://www.unomaha.edu/college-of-information-science-and-technology/natural-language-processing-and-knowledge-representation-lab/_files/papers/Text2Drses_system_description.pdf}}.

\bibitemdeclare{article}{ManningSBFBM14}
\bibitem{ManningSBFBM14}
\bibinfo{author}{Christopher~D \surnamestart Manning\surnameend},
  \bibinfo{author}{Mihai \surnamestart Surdeanu\surnameend},
  \bibinfo{author}{John \surnamestart Bauer\surnameend}, \bibinfo{author}{Jenny
  \surnamestart Finkel\surnameend}, \bibinfo{author}{Steven \surnamestart
  Bethard\surnameend} \& \bibinfo{author}{David \surnamestart
  McClosky\surnameend} (\bibinfo{year}{2014}): \emph{\bibinfo{title}{{The
  Stanford CoreNLP Natural Language Processing Toolkit}}}.
\newblock {\sl \bibinfo{journal}{Proceedings of 52nd Annual Meeting of the
  Association for Computational Linguistics: System Demonstrations}}, pp.
  \bibinfo{pages}{55--60}, \doi{10.3115/v1/P14-5010}.
\newblock \urlprefix\url{http://aclweb.org/anthology/P14-5010}.

\bibitemdeclare{article}{MishraDHTYC18}
\bibitem{MishraDHTYC18}
\bibinfo{author}{Bhavana~Dalvi \surnamestart Mishra\surnameend},
  \bibinfo{author}{Lifu \surnamestart Huang\surnameend}, \bibinfo{author}{Niket
  \surnamestart Tandon\surnameend}, \bibinfo{author}{Wen{-}tau \surnamestart
  Yih\surnameend} \& \bibinfo{author}{Peter \surnamestart Clark\surnameend}
  (\bibinfo{year}{2018}): \emph{\bibinfo{title}{Tracking State Changes in
  Procedural Text: {A} Challenge Dataset and Models for Process Paragraph
  Comprehension}}.
\newblock {\sl \bibinfo{journal}{CoRR}} \bibinfo{volume}{abs/1805.06975}.
\newblock \urlprefix\url{http://arxiv.org/abs/1805.06975}.

\bibitemdeclare{inproceedings}{MitraB16}
\bibitem{MitraB16}
\bibinfo{author}{Arindam \surnamestart Mitra\surnameend} \&
  \bibinfo{author}{Chitta \surnamestart Baral\surnameend}
  (\bibinfo{year}{2016}): \emph{\bibinfo{title}{{Addressing a Question
  Answering Challenge by Combining Statistical Methods with Inductive Rule
  Learning and Reasoning}}}.
\newblock In: {\sl
  \bibinfo{booktitle}{{AAAI} Conference on Artificial Intelligence}},
  \bibinfo{publisher}{{AAAI} Press}, pp. \bibinfo{pages}{2779-2785}.

\bibitemdeclare{mastersthesis}{Olson19}
\bibitem{Olson19}
\bibinfo{author}{Craig \surnamestart Olson\surnameend} (\bibinfo{year}{2019}):
  \emph{\bibinfo{title}{{Processing Narratives by Means of Action Languages}}}.
\newblock Master's thesis, \bibinfo{school}{University of Nebraska Omaha}.

\bibitemdeclare{misc}{Palmer18}
\bibitem{Palmer18}
\bibinfo{author}{Martha \surnamestart Palmer\surnameend}
  (\bibinfo{year}{2018}): \emph{\bibinfo{title}{{VerbNet}}}.
\newblock
  \bibinfo{howpublished}{\url{https://verbs.colorado.edu/verb-index/vn3.3/}}.

\bibitemdeclare{article}{propbank}
\bibitem{propbank}
\bibinfo{author}{Martha \surnamestart Palmer\surnameend},
  \bibinfo{author}{Daniel \surnamestart Gildea\surnameend} \&
  \bibinfo{author}{Paul \surnamestart Kingsbury\surnameend}
  (\bibinfo{year}{2005}): \emph{\bibinfo{title}{The Proposition Bank: An
  Annotated Corpus of Semantic Roles}}.
\newblock {\sl \bibinfo{journal}{Computational Linguistics}}
  \bibinfo{volume}{31}(\bibinfo{number}{1}), pp. \bibinfo{pages}{71--106},
  \doi{10.1162/0891201053630264}.

\bibitemdeclare{misc}{WertzCZ18}
\bibitem{WertzCZ18}
\bibinfo{author}{Edward \surnamestart Wertz\surnameend},
  \bibinfo{author}{Anuradha \surnamestart Chandrasekan\surnameend} \&
  \bibinfo{author}{Yuanlin \surnamestart Zhang\surnameend}
  (\bibinfo{year}{2018}): \emph{\bibinfo{title}{{CALM: a Compiler for Modular
  Action Language ALM}}}.
\newblock \bibinfo{howpublished}{unpublished draft}.

\bibitemdeclare{article}{WestonBCM15}
\bibitem{WestonBCM15}
\bibinfo{author}{Jason \surnamestart Weston\surnameend},
  \bibinfo{author}{Antoine \surnamestart Bordes\surnameend},
  \bibinfo{author}{Sumit \surnamestart Chopra\surnameend} \&
  \bibinfo{author}{Tomas \surnamestart Mikolov\surnameend}
  (\bibinfo{year}{2015}): \emph{\bibinfo{title}{Towards {AI}-Complete Question
  Answering: {A} Set of Prerequisite Toy Tasks}}.
\newblock {\sl \bibinfo{journal}{CoRR}} \bibinfo{volume}{abs/1502.05698}.
\newblock \urlprefix\url{http://arxiv.org/abs/1502.05698}.

\end{thebibliography}

\end{document}